# A Deep Framework for Cross-Domain and Cross-System Recommendations

**Feng Zhu**[1], **Yan Wang**[1], **Chaochao Chen**[2], **Guanfeng Liu**[1], **Mehmet Orgun**[1], **Jia Wu**[1]

[1] Department of Computing, Macquarie University, Sydney, NSW 2109, Australia
[2] AI Department, Ant Financial Services Group, Hangzhou 310012, China
feng.zhu3@students.mq.edu.au, yan.wang@mq.edu.au, chaochao.ccc@antfin.com,
guanfeng.liu@mq.edu.au, mehmet.orgun@mq.edu.au, jia.wu@mq.edu.au

## Abstract

Cross-Domain Recommendation (CDR) and Cross-System Recommendations (CSR) are two of the promising solutions to address the long-standing data sparsity problem in recommender systems. They leverage the relatively richer information, e.g., ratings, from the source domain or system to improve the recommendation accuracy in the target domain or system. Therefore, finding an accurate mapping of the latent factors across domains or systems is crucial to enhancing recommendation accuracy. However, this is a very challenging task because of the complex relationships between the latent factors of the source and target domains or systems. To this end, in this paper, we propose a Deep framework for both Cross-Domain and Cross-System Recommendations, called DCDCSR, based on Matrix Factorization (MF) models and a fully connected Deep Neural Network (DNN). Specifically, DCDCSR first employs the MF models to generate user and item latent factors and then employs the DNN to map the latent factors across domains or systems. More importantly, we take into account the rating sparsity degrees of individual users and items in different domains or systems and use them to guide the DNN training process for utilizing the rating data more effectively. Extensive experiments conducted on three real-world datasets demonstrate that DCDCSR framework outperforms the state-of-the-art CDR and CSR approaches in terms of recommendation accuracy.

## 1 Introduction

Data sparsity is a long-standing problem in recommender systems (RSs). In order to address this problem, a new trend has emerged in recent years by making the use of the relatively richer information, e.g., ratings, from the source domain or system to improve the recommendation accuracy in the target domain or system. Such approaches are referred to as *Cross-Domain Recommendation* (CDR) [Berkovsky *et al.*, 2007] and *Cross-System Recommendation* (CSR) [Zhao *et al.*, 2013], respectively. For example, on Douban website (https://www.douban.com), its recommender system can recommend books to users according to their movie reviews (i.e., CDR) since users in different domains are very likely to have similar tastes. In addition, the movie features derived from Netflix system can be transferred to Douban system [Zhao *et al.*, 2013] (i.e., CSR) because both Netflix and Douban have the same domain of movie reviews.

The existing CDR approaches can be classified into two groups, i.e., content-based approaches and transfer-based approaches. *Content-based Approaches* in CDR tend to link different domains by identifying similar user/item attributes [Chung *et al.*, 2007], social tags [Fernández-Tobías and Cantador, 2014], and user-generated texts [Tan *et al.*, 2014]. In contrast, *Transfer-based Approaches* in CDR mainly focus on transferring latent factors [Pan *et al.*, 2010] or rating patterns [Gao *et al.*, 2013] from the source domain to the target domain. Different from the content-based approaches, transfer-based approaches typically employ machine learning techniques, such as transfer learning [Zhang *et al.*, 2016] and neural networks [Man *et al.*, 2017], to transfer knowledge across domains. Like CDR, Cross-System Recommendation (CSR) is also an effective solution for the data sparsity problem. CSR leverages the ratings or the knowledge derived from the source system to improve the recommendation accuracy in the target system, where both systems are in the same domain [Zhao *et al.*, 2013].

The common idea of the existing transfer-based approaches in CDR and CSR is to map the latent factors obtained from a source domain or system (a relatively richer data source) to a target domain or system (a sparser data source) for improving the recommendation accuracy. Therefore, accurately mapping the latent factors across domains or systems is crucial for enhancing the recommendation accuracy in CDR and CSR. However, the existing transfer-based approaches cannot effectively obtain an accurate mapping between the latent factors in two domains or systems. They either directly replace a part of the latent factors in the target domain or system with the corresponding latent factors in the source domain or system [Zhao *et al.*, 2017] (*Category 1*), or map the latent factors of common users/items in the source domain or system to fit those in the target domain or system [Man *et al.*, 2017] (*Category 2*). The approaches in *Category 1* ignore the complex relationship between the latent factors in the two domains or systems, while the approaches in *Cat-* *egory 2* only focus on the common users and items to adjust their relatively accurate latent factors in the source domain or system to fit the worse ones in the target domain or system, which is not reasonable and effective.

**Our Approach and Contributions:** Different from the existing CDR and CSR approaches, we propose a novel approach to generating benchmark factors, which combine the features of the latent factors in both the source and the target domains or systems. We then map the latent factors in the target domain or system to fit the benchmark factors. To the best of our knowledge, this leads to a new category of transfer-based approaches to mapping latent factors across domains or systems, and our approach is the first one in this novel category.

The characteristics and contributions of our framework are summarized as follows:
(1) In this paper, we propose a Deep framework for both Cross-Domain and Cross-System Recommendations, called DCDCSR, which employs MF models and a fully connected Deep Neural Network (DNN);
(2) We employ the MF models to generate user and item latent factors. When generating benchmark factors, we take into account fine-grained sparsity degrees of individual users and items to combine the latent factors learned from both the source and target domains or systems, which can effectively utilize more rating data in the two domains or systems;
(3) We employ the DNN to accurately map the latent factors in the target domain or system to fit the benchmark factors, which can improve recommendation accuracy;
(4) The extensive experiments conducted on three real-world datasets demonstrate that our DCDCSR framework outperforms the state-of-the-art approaches, which clearly improves the recommendation accuracy for both CDR and CSR.

## 2 Related Work

In this section, we review the existing CDR approaches in two groups: (1) content-based approaches and (2) transfer-based approaches, and the existing works on CSR.

### 2.1 Content-Based Approaches for CDR

*Cross-Domain Recommendation* (CDR) was first proposed in [Berkovsky *et al.*, 2007], which is a content-based approach targeting the data sparsity problem by merging user preferences and extracting common attributes of users and items. Later on, the work proposed in [Winoto and Tang, 2008] uncovers the relationships of user preferences in different domains, and the work proposed in [Berkovsky *et al.*, 2008] imports and integrates the data collected by other systems to acquire accurate modeling of users' interests and needs. In addition to considering user/item attributes, linking domains by other information is also a typical solution in this group, such as social tags [Fernández-Tobías and Cantador, 2014], and user-generated texts [Tan *et al.*, 2014].

### 2.2 Transfer-Based Approaches for CDR

Transfer-based approaches mainly employ MF models to generate latent factors or rating patterns and transfer them across domains. The work proposed in [Singh and Gordon, 2008] suggests a non-linear relationship to share the latent factors of entities across domains. Later on, the work proposed in [Pan *et al.*, 2011] utilizes a matrix-based transfer learning framework to combine both user and item knowledge in source domains. In addition to transferring latent factors, transferring rating patterns [Li *et al.*, 2009] becomes effective in transfer-based approaches. In [Agarwal *et al.*, 2011], each entity shares a global latent factor generated by a domain-specific transfer matrix, which means that this factor can be used in both the source and the target domains directly.

### 2.3 Cross-System Recommendation (CSR)

*Cross-System Recommendation* (CSR) emerged later than CDR. The first work for CSR was proposed in [Zhao *et al.*, 2013] and improved in [Zhao *et al.*, 2017] by employing transfer learning techniques to recommend unrated items across systems. Recently, the EMCDR framework proposed in [Man *et al.*, 2017] supports both CDR and CSR but can only utilize common items or common users as a bridge.

**Summary:** For CDR, while the existing content-based approaches have difficulties in obtaining more user profiles and item details [Lops *et al.*, 2011], the existing transfer-based approaches require a certain level of overlap between two domains, e.g., common users or items, which restricts them from fully utilizing historical feedback data to improve recommendation accuracy. For CSR, the existing approaches either directly utilize the latent factors learned from the source system [Zhao *et al.*, 2017] or take some unreasonable mapping strategies [Man *et al.*, 2017], each of which leads to a low recommendation accuracy.

## 3 The Proposed DCDCSR Framework

In this section, we first formulate the Cross-Domain and Cross-System Recommendation problems. Then, we propose a Deep framework for both Cross-Domain and Cross-System Recommendations, called DCDCSR, and introduce our DNN mapping process for mapping latent factors across domains or systems. We also introduce how to make cross-domain and cross-system recommendations based on the predicted ratings.

### 3.1 Notations and Problem Definition

Let $\boldsymbol{R}^s$ and $\boldsymbol{R}^t$ denote the rating matrices of the source and target domains or systems, respectively. Let $\mathcal{U} = \{u_1, ..., u_n\}$ and $\mathcal{V} = \{v_1, ..., v_m\}$ denote the user and item sets, respectively, where $n$ is the number of users and $m$ is the number of items. $r_{ij}^t \in \boldsymbol{R}^t$ denotes the rating that $u_i$ gives to item $v_j$ in the target domain or system. Given a rating matrix $\boldsymbol{R}$, after matrix factorization, $\boldsymbol{R}$ is factorized into two latent matrices $\boldsymbol{U}$ ($K \times n$) and $\boldsymbol{V}$ ($K \times m$), where $K$ is the dimension of factors. $\boldsymbol{U}$ and $\boldsymbol{V}$ represent the low-rank factor matrices for $\mathcal{U}$ and $\mathcal{V}$, respectively. Concretely, $\boldsymbol{U}_i^t$ denotes $u_i$'s latent factor vector in the target domain or system. Based on these notations, the Cross-Domain Recommendation (CDR) problem can be defined as follows.

**Definition 1. Cross-Domain Recommendation (CDR) Problem:** *Input: Two observed domains including the rating matrices $\boldsymbol{R}^s$ and $\boldsymbol{R}^t$, the user sets $\mathcal{U}^s, \mathcal{U}^t \subseteq \mathcal{U}$, and the item sets $\mathcal{V}^s, \mathcal{V}^t \subseteq \mathcal{V}$. Output: Recommend the items*

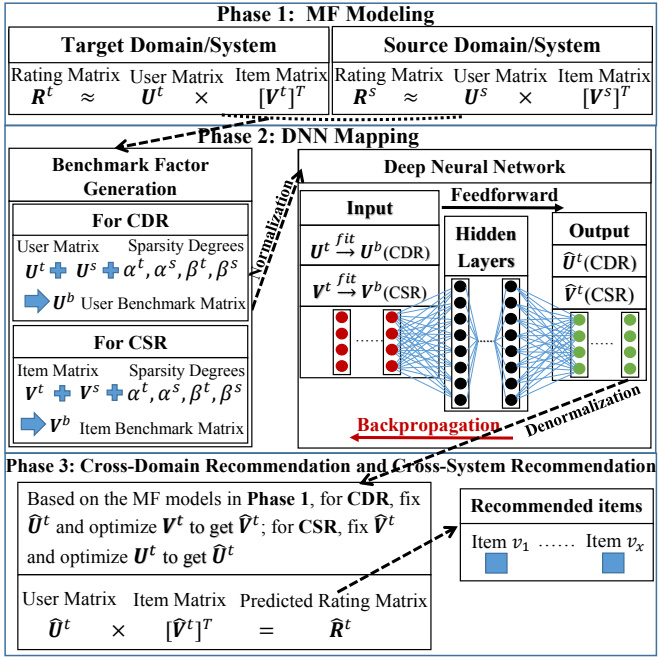

Figure 1: The Structure of our DCDCSR Framework

**Algorithm 1** The DCDCSR Framework

**Require:** The rating matrices, user sets, and item sets of the source and target domains or systems $R^s, R^t, \mathcal{U}^s, \mathcal{U}^t, \mathcal{V}^s,$ and $\mathcal{V}^t$.
**Ensure:** Recommend items $\mathcal{V}_i \subseteq \mathcal{V}^t$ to a target user $u_i$ in the target domain or system.
  **Phase 1: MF Modeling**
1: Learn $\{U^s, V^s\}$ from $R^s$ by using matrix factorization;
2: Learn $\{U^t, V^t\}$ from $R^t$ by using matrix factorization.
  **Phase 2: DNN Mapping**
3: Generate the benchmark factor matrix $U^b$ for CDR or $V^b$ for CSR.
4: Normalize $\{U^t, U^b\}$ for CDR or $\{V^t, V^b\}$ for CSR.
5: Train the parameters of the deep neural network by the Feedforward and Backpropagation processes.
6: Obtain the affine factor matrices $\hat{U}^t$ or $\hat{V}^t$.
7: Denormalize $\hat{U}^t$ or $\hat{V}^t$.
  **Phase 3: Cross-Domain Recommendation and Cross-System Recommendation**
8: For CDR, fix $\hat{U}^t$ and train $\hat{V}^t$ from $R^t$ by using the MF model in Phase 1.
9: For CSR, fix $\hat{V}^t$ and train $\hat{U}^t$ from $R^t$ by using the MF model in Phase 1.
10: Obtain the predicted ratings $\hat{R}^t = \hat{U}^t[\hat{V}^t]^\top$ for the target domain or system.
11: **return** $\mathcal{V}_i$.

$\mathcal{V}_i \subseteq \mathcal{V}^t$ *to a target user* $u_i \in \mathcal{U}^t$ *by utilizing both* $R^s$ *and* $R^t$.

Likewise, we can formulate the CSR problem by replacing "domain" with "system" in Definition 1.

### 3.2 The DCDCSR Framework

Targeting the above CDR and CSR problems, we propose a deep framework, called DCDCSR, for both CDR and CSR. This framework can be divided into three phases, i.e., **Phase 1**: MF Modeling, **Phase 2**: DNN Mapping, and **Phase 3**: Cross-Domain and Cross-System Recommendations. The framework structure is shown in Figure 1.

In Phase 1, we obtain the user and item latent factor matrices $\{U^s, U^t, V^s, V^t\}$ by using matrix factorization. In Phase 2, we first generate the benchmark factor matrices $\{U^b, V^b\}$ by combining the latent factor matrices $\{U^s, U^t, V^s, V^t\}$ according to the sparsity degrees of individual users and items. Then, we train the deep neural network in the *Feedforward* and the *Backpropagation* processes to map the latent factor matrices $\{U^t, V^t\}$ to fit the benchmark factor matrices $\{U^b, V^b\}$. In Phase 3, based on the affine factor matrices $\{\hat{U}^t, \hat{V}^t\}$ learned from Phase 2, we predict the users' ratings on all items in the target domain or system and recommend matched items to target users. The three phases in the framework are presented in Algorithm 1 with details explained in the following sections.

### 3.3 Phase 1: MF Modeling

To study the generalizability of our proposed DCDCSR framework, in this Phase, we apply two classical rating-oriented MF models (MMMF and PMF) and a representative ranking-oriented MF model (BPR) to generate user and item latent factors for the following mapping process. While the rating-oriented MF models focus on minimizing the error between observed and predicted ratings, the ranking-oriented MF model emphasizes to remain the personalized rating rankings on items unchanged between observed and predicted ratings, all of which can bring different biased latent factors into the following DNN mapping process. Due to space constraints, we only briefly introduce them below.

**Rating-Oriented Matrix Factorization**

*Maximum-Margin Matrix Factorization* (MMMF) [Srebro et al., 2005] learns a matrix $\hat{R}$ to fit the observed rating matrix $R$ by minimizing a trace norm of $R$ and maximizing the corresponding predictive margin.

*Probabilistic Matrix Factorization* (PMF) [Mnih and Salakhutdinov, 2008] is a probabilistic model with Gaussian observation noise and its core idea is to maximize the conditional distribution over the observed ratings.

**Ranking-Oriented Matrix Factorization**

*Bayesian Personalized Ranking model* (BPR) [Rendle et al., 2009] is a generic optimization benchmark for personalized ranking and its core idea is to minimize the ranking error between predicted and observed ratings.

### 3.4 Phase 2: The DNN Mapping

The user and item latent factor matrices $\{U^s, U^t, V^s, V^t\}$ can be learned by the above-mentioned MF models. Next, we develop a fully connected deep neural network to represent the relationship of latent factors between two domains or two systems, i.e., DNN Mapping.

As mentioned in Section 1, both enforcing $\{U^t, V^t\}$ to be the same as $\{U^s, V^s\}$ [Zhao et al., 2017] and mapping $\{U^s, V^s\}$ to fit $\{U^t, V^t\}$ [Man et al., 2017] are not effective and reasonable because the accuracies of user and item latent factors mainly depend on their sparsity degrees. More importantly, a common entity in the source domain or system may be sparser than the one in the target domain or system. This means the latent factors of this entity in the source domain or system are less accurate than the ones in the target domain and system. Therefore, we generate more reasonable benchmark factor matrices $U^b$ and $V^b$ by integrating the latent factors and considering the sparsity degrees of individual users and items in both the source and target domains or systems.

### The Generation of Benchmark Factors

*First*, we extract the *common users* $\mathcal{CU}$ from two different domains for CDR and the *common items* $\mathcal{CV}$ from two different systems for CSR.

*Then*, we define the sparsity degrees of common entities (either users or items) in different domains or systems below.

**Definition 2. Sparsity Degrees of Common Entities:** *For any common entity $e_i \in \mathcal{CU} \cup \mathcal{CV}$, given the total numbers of ratings of $e_i$ in the source and target domains or systems, $n_i^s$ and $n_i^t$, the sparsity degrees $\alpha_i^s$ and $\alpha_i^t$ of the entity $e_i$ in the source and target domains or systems are calculated as*

$$\alpha_i^s = \frac{n_i^t}{(n_i^s + n_i^t)}, \quad \alpha_i^t = 1 - \alpha_i^s. \quad (1)$$

In order to generate more reasonable benchmark factors for the following DNN mapping process, both $\{U^s, V^s\}$ and $\{U^t, V^t\}$ are nonnegligible factors. Thus, based on the idea of feature combination introduced in [Burke, 2002], for each common user $u_i \in \mathcal{CU}$, the benchmark factor vector $CU_i^b$ can be calculated as follows:

$$CU_i^b = (1 - \alpha_i^s) \cdot U_i^s + (1 - \alpha_i^t) \cdot U_i^t. \quad (2)$$

According to Eq.(2), the smaller the sparsity degrees $\alpha_i^s$ and $\alpha_i^t$, the more accurate their corresponding latent factor vectors $U_i^s$ and $U_i^t$, and hence the more we take these vectors into account in generating benchmark factor vector $CU_i^b$.

Likewise, for CSR, we can obtain the benchmark factor matrix $CV^b$ for the common items.

*Next*, we find out the *different users* $\mathcal{DU}^t = \mathcal{U}^t - \mathcal{CU}$ in the target domain for CDR and the *different items* $\mathcal{DV}^t = \mathcal{V}^t - \mathcal{CV}$ in the target system for CSR. Later on, we employ the cosine similarity to measure the similarities between common entities and different entities. For each user $u_i \in \mathcal{DU}^t$, we choose the top-$k$ similar users $\mathcal{SU}_i$ from $\mathcal{CU}$. Similarly, for each item $v_i \in \mathcal{DV}^t$, we also choose the top-$k$ similar items $\mathcal{SV}_i$ from $\mathcal{CV}$. Based on these top-$k$ similar entities, we define the sparsity degrees of different entities as follows.

**Definition 3. Sparsity Degrees of Different Entities:** *For any different entity $e_i \in \mathcal{DU}^t \cup \mathcal{DV}^t$, given the total number of ratings of $e_i$ in the target domain or system, $n_i^s$, and the average number of ratings of $e_i$'s top-$k$ similar entities in the source domain or system, $sn_i^s$, the sparsity degree $\beta_i^t$ of entity $e_i$ in the target domain or system is calculated as*

$$\beta_i^t = \frac{sn_i^s}{(n_i^t + sn_i^s)}. \quad (3)$$

Thus, for each different user $u_i \in \mathcal{DU}^t$, the benchmark factor vector $DU_i^b$ can be calculated as

$$DU_i^b = (1 - \beta_i^t) \cdot U_i^t + \beta_i^t \cdot SU_i, \text{where}$$
$$SU_i = \frac{\sum_{u_j \in \mathcal{SU}_i} sim(u_i, u_j) \cdot U_j^s}{\sum_{u_j \in \mathcal{SU}_i} sim(u_i, u_j)}. \quad (4)$$

Similarly, for CSR, we can obtain the benchmark factor matrix $DV^b$ for different items.

*Finally*, We have $U^b = CU^b \cup DU^b$ and $V^b = CV^b \cup DV^b$.

### The Mapping Process

**Normalization:** The benchmark factor matrices $\{U^b, V^b\}$ can be obtained by the above-mentioned feature combination method. We first normalize the latent factor matrices $\{U^t, V^t\}$ and the benchmark factor matrices $\{U^b, V^b\}$ into the range $[-1, 1]$ by using the mapminmax function.

**Mapping Process:** As shown in Phase 2 of Figure 1, we employ a fully connected deep neural network to map $U^t$ to fit $U^b$ for CDR and map $V^t$ to fit $V^b$ for CSR, respectively. Since the mapping processes for CDR and CSR are similar, we take CDR as an example to introduce the DNN mapping process. In general, in order to minimize the mapping loss for CDR, the process of training mapping parameters can be changed into the following minimization problem,

$$\min_{\Theta} \ell(h(U^t; \Theta), U^b), \quad (5)$$

where $h(\cdot)$ is a DNN mapping function introduced below and the loss function $\ell(\cdot)$ is the square loss.

The detailed training process is divided into two steps, i.e., Feedforward and Backpropagation.

- **Feedforward:** Each latent factor vector is denoted as a low-dimensional vector, and in each layer, each input vector is mapped into a hidden vector. Let $x_j$ denote the input vector, $W_j$ denote the weight vector, $b_j$ denote the bias term, and $y_j$ denote the output vector for the $j$-th hidden layer, $j = 1, ..., d$. Thus, for $U_i^t \subset U^t$, we have

$$\begin{aligned} x_1 &= U_i^t, \\ y_j &= f(W_j \cdot x_j + b_j), \quad j = 1, ..., d-1, \\ h(U_i^t; \Theta) &= f(W_d \cdot x_d + b_d), \quad \Theta = \{W; b\}, \end{aligned} \quad (6)$$

  and we choose the tan-sigmoid function as the activation function, i.e., $f(x) = \frac{2}{(1+\exp{-2x})} - 1$.

- **Backpropagation:** According to the chain rule, we recursively update the parameters by computing the gradients of all inputs, parameters, and intermediates as introduced in [Riedmiller and Braun, 1993].

For CSR, the mapping process is similar, and we just replace $\{U^t, U^b\}$ with $\{V^t, V^b\}$ as the input for the DNN.

**Denormalization:** After the DNN mapping, we obtain the affine factor matrix $\hat{U}^t$ for CDR and $\hat{V}^t$ for CSR, respectively. Finally, we denormalize the affine factor matrix into the range of the original latent factor matrix by reversing the mapminmax function.

### 3.5 Phase 3: Cross-Domain Recommendation and Cross-System Recommendation

**CDR:** The items in the source and target domains are definitely different. Thus, for CDR, with the DNN mapping, we can obtain the affine factor matrix $\hat{U}^t$. However, original $V^t$ has yet to be improved. To this end, we fix $\hat{U}^t$ and only update $V^t$ by using the MF models to generate $\hat{V}^t$.

**CSR:** Similarly, either the users in the source and target systems are totally different or it is difficult to determine whether they are the same. Thus, for CSR, we first obtain $\hat{V}^t$ by the DNN mapping, then fix $\hat{V}^t$ and obtain $\hat{U}^t$ by using the MF models.

| Tasks | Cross-Domain | | | Cross-System | | |
|---|---|---|---|---|---|---|
| Datasets | Douban | | | Netflix | MovieLens | Douban* |
| Domains | Movie | Book | Music | Movie | Movie | Movie |
| #Users | 3,982 | 3,032 | 1,983 | 59,688 | 138,493 | 500 |
| #Items | 90,553 | 87,848 | 88,986 | 17,434 | 27,278 | 90,553 |
| #Ratings | 2,326,913 | 239,330 | 242,013 | 2,000,000 | 20,000,263 | 48,619 |

Table 1: Experimental datasets

Finally, based on $\hat{U}^t$ and $\hat{V}^t$, we can recommend matched items $\mathcal{V}_i \subseteq \mathcal{V}^t$ to target user $u_i \in \mathcal{U}^t$ for both CDR and CSR.

## 4 Experiments and Analysis

Extensive experiments are conducted on three real-world datasets, which aim to answer the following questions:
**Q1:** How does the dimension $K$ of the latent factors affect the efficiency of our DCDCSR framework? (in Result 1)
**Q2:** How does our approach outperform the state-of-the-art approaches for both Cross-Domain and Cross-System Recommendations? (in Results 2 & 3)

### 4.1 Experimental Settings

**Datasets:** In the experiments, we use three real-world datasets, namely two public benchmark datasets Netflix Prize[1] and MovieLens 20M[2], and a Douban dataset crawled from the Douban website. Since MovieLens 20M contains more than 20 million ratings (relatively richer), for the diversity of our experiments, we extract a subset from Netflix Prize, with a smaller scale of ratings (2 million). The details of these three datasets are shown in Table 1.

For the CDR experiments, we take DoubanMovie as the source domain corresponding to the target domains DoubanBook and DoubanMusic. For the CSR experiments, we take Netflix and MovieLens as the source systems and extract a subset Douban*Movie from DoubanMovie as the target system. The numbers of common items of Netflix-Douban* and MovieLens-Douban* are 3,700 and 5,712, respectively. For the Douban dataset, the numbers of common users of Movie-Music and Movie-Book are 295 and 379, respectively.

In our experiments, we split each dataset into a training set (80%) with the early ratings and a test set (20%) with the later ratings. The sequences of ratings and latent factors may slightly affect the performances of matrix factorization and mapping respectively. Thus, we report the average results of 5 random times.

**Parameter Setting:** We set the dimension $K$ of the latent factor as 10, 20, 50, and 100, respectively. In order to generate the benchmark factors, we set $k = 5$ for top-$k$ similar items or users. For the deep neural network, we set the depth of the hidden layers $d$ to 5 because when $d > 5$, the performances of our methods almost do not change. We set the dimension of the input and output of the DNN to $K$, and the number of hidden nodes to $1.5 \times K$. We randomly initialize the parameters as suggested in [Glorot and Bengio, 2010], i.e., $W \sim U[-\frac{1}{\sqrt{2K}}, \frac{1}{\sqrt{2K}}]$. In addition, we set the batch size to 32, and the learning rate to 0.005.

**Experimental Tasks and Evaluation Metrics:** In total, we design two CDR tasks and two CSR tasks as follows:

[1] https://www.kaggle.com/netflix-inc/netflix-prize-data
[2] https://www.kaggle.com/grouplens/movielens-20m-dataset

**Task 1:** DoubanMovie → DoubanBook (for CDR),
**Task 2:** DoubanMovie → DoubanMusic (for CDR),
**Task 3:** Netflix → Douban*Movie (for CSR),
**Task 4:** MovieLens → Douban*Movie (for CSR).

We use the Mean Absolute Error (MAE) and the Root Mean Squared Error (RMSE) as metrics to evaluate recommendation performance, which are commonly used in the literature for CDR and CSR [Pan *et al.*, 2010; Zhao *et al.*, 2017].
**Comparison Methods:** In the experiments, we implement our DCDCSR framework into three methods by applying MMMF, PMF, and BPR as the MF models, i.e., MMMF_DCDCSR, PMF_DCDCSR, and BPR_DCDCSR.

We compare our three DCDCSR methods with the following seven methods implemented from three representative models:
(1) *Bayesian Personalized Ranking model* (BPR) [Rendle *et al.*, 2009]: BPR is a ranking-oriented MF model. We choose it as a conventional baseline method running on the target domain and system, which does not take any Cross-Domain or Cross-System strategies.
(2) *Active transfer learning framework* (ATL) [Zhao *et al.*, 2017]: This is a state-of-the-art framework which utilizes Transfer Learning (TL). It offers three methods. In our experiments, we choose the two well-performing methods MMMF_TL and PMF_TL.
(3) *Embedding and Mapping framework* (EMCDR) [Man *et al.*, 2017]: This is a state-of-the-art framework which utilizes Linear Matrix Translation (LIN) and Multi-Layer Perceptron (MLP). It adopts PMF and BPR as its MF models, and maps the latent factors across domains or systems with both LIN and MLP ($2 \times 2$). Thus, this framework offers four methods, namely, MF_EMCDR_LIN, MF_EMCDR_MLP, BPR_EMCDR_LIN and BPR_EMCDR_MLP, all of which are compared in our experiments.

### 4.2 Performance Comparison and Analysis

All the experimental results are presented in Table 2.

**Result 1: Impact of Latent Factor Dimension**
In order to answer question **Q1**, we investigate how the performance of DCDCSR framework is affected by the dimension $K$ of the latent factors. From Table 2, we can see that when $K = 10$ or 20, in general, the performances of DCDCSR methods increase (i.e., the MAE and RMSE decrease) with $K$. However, when $K = 50$, there is no significant improvement in the performance. Moreover, when $K = 100$, the performances have a slight decline. This is because the number of parameters of the DNN geometrically increases with $K$. When $K = 100$, while the training data remains the same, the performance of the DNN mapping declines slightly.

**Result 2: Cross-Domain Recommendation (Tasks 1 & 2)**
In order to answer question **Q2**, we compare the performances of our methods and the seven comparison methods in the CDR tasks (Tasks 1 & 2). From Table 2, we can see that, for the CDR tasks, MMMF_DCDCSR does not perform as well as PMF_DCDCSR and BPR_DCDCSR because its MF model cannot effectively learn a predicted matrix $\hat{R}$ by maximizing the predictive trace margin on the target domains DoubanBook and DoubanMusic. In terms of

|  |  | Cross-Domain Recommendation (CDR) | | | | Cross-System Recommendation (CSR) | | | |
|---|---|---|---|---|---|---|---|---|---|
|  |  | Task 1 | | Task 2 | | Task 3 | | Task 4 | |
|  |  | MAE | RMSE | MAE | RMSE | MAE | RMSE | MAE | RMSE |
| K=10 | BPR | 0.7187 (± 0.0011) | 0.9386 (± 0.0014) | 0.7231 (± 0.0012) | 0.9416 (± 0.0017) | 0.7524 (± 0.0014) | 0.9628 (± 0.0016) | 0.7524 (± 0.0014) | 0.9628 (± 0.0016) |
|  | MMMF_TL | 0.7001 (± 0.0009) | 0.9128 (± 0.0007) | 0.6978 (± 0.0006) | 0.9093 (± 0.0005) | 0.7162 (± 0.0012) | 0.8951 (± 0.0003) | 0.7090 (± 0.0007) | 0.8997 (± 0.0003) |
|  | PMF_TL | 0.7022 (± 0.0016) | 0.9187 (± 0.0006) | 0.7077 (± 0.0008) | 0.9097 (± 0.0005) | 0.7031 (± 0.0008) | 0.8913 (± 0.0012) | 0.7120 (± 0.0003) | 0.9030 (± 0.0007) |
|  | MF_EMCDR_LIN | 0.7065 (± 0.0003) | 0.9103 (± 0.0006) | 0.7024 (± 0.0012) | 0.9163 (± 0.0004) | 0.7096 (± 0.0008) | 0.9113 (± 0.0007) | 0.7340 (± 0.0009) | 0.9326 (± 0.0007) |
|  | MF_EMCDR_MLP | 0.7011 (± 0.0015) | 0.9071 (± 0.0009) | 0.7022 (± 0.0008) | 0.9045 (± 0.0012) | 0.7087 (± 0.0008) | 0.9049 (± 0.0005) | 0.7045 (± 0.0004) | 0.9062 (± 0.0005) |
|  | BPR_EMCDR_LIN | 0.7084 (± 0.0012) | 0.9111 (± 0.0006) | 0.7065 (± 0.0005) | 0.9105 (± 0.0013) | 0.7038 (± 0.0004) | 0.9035 (± 0.0003) | 0.7080 (± 0.0005) | 0.9043 (± 0.0006) |
|  | BPR_EMCDR_MLP | 0.7061 (± 0.0005) | 0.9054 (± 0.0005) | 0.6987 (± 0.0003) | 0.9055 (± 0.0008) | 0.6995 (± 0.0005) | 0.8994 (± 0.0003) | 0.6991 (± 0.0002) | 0.8994 (± 0.0005) |
|  | **MMMF_DCDCSR** | 0.7041 (± 0.0005) | 0.8971 (± 0.0004) | 0.6992 (± 0.0003) | 0.8875 (± 0.0002) | 0.6998 (± 0.0003) | 0.8865 (± 0.0002) | 0.6994 (± 0.0005) | 0.8836 (± 0.0004) |
|  | **PMF_DCDCSR** | 0.7037 (± 0.0005) | 0.8965 (± 0.0003) | 0.6996 (± 0.0004) | **0.8866 (± 0.0002)** | 0.6838 (± 0.0012) | 0.8681 (± 0.0011) | **0.6753 (± 0.0006)** | **0.8659 (± 0.0007)** |
|  | **BPR_DCDCSR** | **0.6943 (± 0.0003)** | **0.8881 (± 0.0006)** | **0.6971 (± 0.0008)** | 0.8872 (± 0.0004) | **0.6786 (± 0.0007)** | **0.8651 (± 0.0008)** | 0.6854 (± 0.0014) | 0.8712 (± 0.0009) |
| K=20 | BPR | 0.7146 (± 0.0014) | 0.9292 (± 0.0007) | 0.7234 (± 0.0011) | 0.9104 (± 0.0006) | 0.7432 (± 0.0012) | 0.9532 (± 0.0014) | 0.7432 (± 0.0012) | 0.9532 (± 0.0014) |
|  | MMMF_TL | 0.7068 (± 0.0004) | 0.9146 (± 0.0002) | 0.7109 (± 0.0003) | 0.9104 (± 0.0002) | 0.6915 (± 0.0002) | 0.8922 (± 0.0002) | 0.7026 (± 0.0003) | 0.8986 (± 0.0002) |
|  | PMF_TL | 0.7017 (± 0.0003) | 0.9188 (± 0.0008) | 0.7176 (± 0.0004) | 0.9244 (± 0.0006) | 0.7024 (± 0.0003) | 0.8969 (± 0.0002) | 0.7057 (± 0.0003) | 0.9012 (± 0.0003) |
|  | MF_EMCDR_LIN | 0.7015 (± 0.0008) | 0.9070 (± 0.0006) | 0.7021 (± 0.0006) | 0.9076 (± 0.0019) | 0.7027 (± 0.0005) | 0.9074 (± 0.0013) | 0.6977 (± 0.0015) | 0.9032 (± 0.0002) |
|  | MF_EMCDR_MLP | 0.7021 (± 0.0005) | 0.9095 (± 0.0005) | 0.7001 (± 0.0003) | 0.9095 (± 0.0004) | 0.6995 (± 0.0003) | 0.8995 (± 0.0002) | 0.6993 (± 0.0005) | 0.8995 (± 0.0005) |
|  | BPR_EMCDR_LIN | 0.7041 (± 0.0009) | 0.9174 (± 0.0005) | 0.7021 (± 0.0008) | 0.9147 (± 0.0012) | 0.7060 (± 0.0007) | 0.9024 (± 0.0005) | 0.6949 (± 0.0006) | 0.9012 (± 0.0008) |
|  | BPR_EMCDR_MLP | 0.7023 (± 0.0008) | 0.9074 (± 0.0005) | 0.7021 (± 0.0008) | 0.9047 (± 0.0012) | 0.6991 (± 0.0005) | 0.8993 (± 0.0003) | 0.6995 (± 0.0002) | 0.8999 (± 0.0002) |
|  | **MMMF_DCDCSR** | 0.7001 (± 0.0002) | 0.8876 (± 0.0004) | 0.6987 (± 0.0003) | 0.8866 (± 0.0003) | 0.7004 (± 0.0003) | 0.8875 (± 0.0004) | 0.7012 (± 0.0001) | 0.8816 (± 0.0004) |
|  | **PMF_DCDCSR** | 0.7003 (± 0.0004) | 0.8872 (± 0.0005) | 0.6985 (± 0.0003) | 0.8879 (± 0.0004) | 0.6880 (± 0.0001) | 0.8609 (± 0.0006) | 0.6805 (± 0.0004) | 0.8654 (± 0.0001) |
|  | **BPR_DCDCSR** | **0.6941 (± 0.0002)** | **0.8845 (± 0.0001)** | **0.6949 (± 0.0004)** | **0.8867 (± 0.0003)** | **0.6723 (± 0.0002)** | **0.8556 (± 0.0008)** | **0.6780 (± 0.0003)** | **0.8601 (± 0.0002)** |
| K=50 | BPR | 0.7115 (± 0.0014) | 0.9413 (± 0.0011) | 0.7252 (± 0.0005) | 0.9464 (± 0.0008) | 0.7252 (± 0.0012) | 0.9364 (± 0.0018) | 0.7252 (± 0.0012) | 0.9364 (± 0.0018) |
|  | MMMF_TL | 0.7062 (± 0.0010) | 0.9189 (± 0.0009) | 0.7143 (± 0.0007) | 0.9132 (± 0.0006) | 0.6899 (± 0.0002) | 0.8851 (± 0.0004) | 0.6948 (± 0.0003) | 0.8975 (± 0.0004) |
|  | PMF_TL | 0.7022 (± 0.0005) | 0.9203 (± 0.0004) | 0.7121 (± 0.0012) | 0.9287 (± 0.0007) | 0.7011 (± 0.0012) | 0.8954 (± 0.0010) | 0.7021 (± 0.0007) | 0.8974 (± 0.0012) |
|  | MF_EMCDR_LIN | 0.7051 (± 0.0003) | 0.9080 (± 0.0002) | 0.7021 (± 0.0008) | 0.9082 (± 0.0005) | 0.7095 (± 0.0014) | 0.9062 (± 0.0005) | 0.7012 (± 0.0007) | 0.9055 (± 0.0009) |
|  | MF_EMCDR_MLP | 0.7065 (± 0.0005) | 0.9114 (± 0.0006) | 0.7076 (± 0.0004) | 0.9086 (± 0.0008) | 0.6997 (± 0.0005) | 0.8997 (± 0.0004) | 0.6993 (± 0.0006) | 0.8995 (± 0.0002) |
|  | BPR_EMCDR_LIN | 0.7055 (± 0.0007) | 0.9086 (± 0.0004) | 0.7020 (± 0.0002) | 0.9084 (± 0.0003) | 0.7013 (± 0.0004) | 0.9034 (± 0.0012) | 0.6983 (± 0.0005) | 0.9016 (± 0.0008) |
|  | BPR_EMCDR_MLP | 0.6917 (± 0.0003) | 0.8994 (± 0.0005) | 0.6987 (± 0.0003) | 0.9003 (± 0.0001) | 0.7022 (± 0.0011) | 0.8981 (± 0.0005) | 0.6995 (± 0.0004) | 0.9003 (± 0.0001) |
|  | **MMMF_DCDCSR** | 0.7003 (± 0.0002) | 0.8880 (± 0.0003) | 0.6988 (± 0.0001) | 0.8889 (± 0.0001) | 0.6924 (± 0.0007) | 0.8856 (± 0.0005) | 0.6935 (± 0.0002) | 0.8746 (± 0.0003) |
|  | **PMF_DCDCSR** | 0.6941 (± 0.0001) | 0.8871 (± 0.0002) | **0.6918 (± 0.0004)** | 0.8925 (± 0.0002) | 0.6820 (± 0.0012) | 0.8655 (± 0.0009) | 0.6794 (± 0.0014) | 0.8636 (± 0.0011) |
|  | **BPR_DCDCSR** | 0.6954 (± 0.0002) | **0.8862 (± 0.0003)** | 0.6957 (± 0.0002) | **0.8874 (± 0.0002)** | **0.6712 (± 0.0008)** | **0.8555 (± 0.0007)** | **0.6595 (± 0.0003)** | **0.8564 (± 0.0002)** |
| K=100 | BPR | 0.7199 (± 0.0005) | 0.9332 (± 0.0011) | 0.7303 (± 0.0005) | 0.9396 (± 0.0005) | 0.7334 (± 0.0012) | 0.9321 (± 0.0004) | 0.7334 (± 0.0012) | 0.9321 (± 0.0004) |
|  | MMMF_TL | 0.7104 (± 0.0003) | 0.9191 (± 0.0002) | 0.7124 (± 0.0003) | 0.9241 (± 0.0001) | 0.6931 (± 0.0002) | 0.8772 (± 0.0002) | 0.6948 (± 0.0003) | 0.8997 (± 0.0002) |
|  | PMF_TL | 0.7089 (± 0.0005) | 0.9213 (± 0.0004) | 0.7071 (± 0.0008) | 0.9207 (± 0.0005) | 0.7020 (± 0.0003) | 0.8966 (± 0.0002) | 0.6995 (± 0.0003) | 0.8954 (± 0.0005) |
|  | MF_EMCDR_LIN | 0.6994 (± 0.0012) | 0.9094 (± 0.0009) | 0.7026 (± 0.0009) | 0.9097 (± 0.0003) | 0.7045 (± 0.0005) | 0.9060 (± 0.0003) | 0.6961 (± 0.0002) | 0.9082 (± 0.0012) |
|  | MF_EMCDR_MLP | 0.7014 (± 0.0002) | 0.9001 (± 0.0004) | 0.7011 (± 0.0004) | 0.8991 (± 0.0005) | 0.7001 (± 0.0002) | 0.9008 (± 0.0008) | 0.6998 (± 0.0012) | 0.9004 (± 0.0001) |
|  | BPR_EMCDR_LIN | 0.6985 (± 0.0004) | 0.9098 (± 0.0001) | 0.7030 (± 0.0008) | 0.9099 (± 0.0003) | 0.7077 (± 0.0011) | 0.9072 (± 0.0002) | 0.7017 (± 0.0006) | 0.9099 (± 0.0008) |
|  | BPR_EMCDR_MLP | 0.7024 (± 0.0006) | 0.8981 (± 0.0003) | 0.7089 (± 0.0001) | 0.8972 (± 0.0002) | 0.6999 (± 0.0005) | 0.9000 (± 0.0001) | 0.6995 (± 0.0002) | 0.9003 (± 0.0006) |
|  | **MMMF_DCDCSR** | 0.7004 (± 0.0002) | 0.8904 (± 0.0002) | 0.7005 (± 0.0002) | 0.8932 (± 0.0003) | 0.6915 (± 0.0004) | 0.8798 (± 0.0002) | 0.6865 (± 0.0009) | 0.8769 (± 0.0004) |
|  | **PMF_DCDCSR** | 0.6986 (± 0.0001) | 0.8895 (± 0.0004) | **0.6942 (± 0.0001)** | 0.8931 (± 0.0001) | 0.6852 (± 0.0012) | 0.8718 (± 0.0009) | 0.6814 (± 0.0003) | 0.8665 (± 0.0004) |
|  | **BPR_DCDCSR** | **0.6971 (± 0.0001)** | **0.8882 (± 0.0002)** | 0.6998 (± 0.0003) | **0.8904 (± 0.0001)** | **0.6745 (± 0.0008)** | **0.8612 (± 0.0011)** | **0.6678 (± 0.0004)** | **0.8594 (± 0.0002)** |

Table 2: The experimental results of CDR and CSR

MAE, PMF_DCDCSR performs the best and it outperforms the seven comparison methods by an average of 1.42%, ranging from 0.94% to 3.57%. Moreover, in terms of RMSE, BPR_DCDCSR performs the best and it outperforms the seven comparison methods by an average of 2.6%, ranging from 1.66% to 5.41%. Compared to all the seven comparison methods, our methods perform clearly better because our sparsity guided DNN mapping process can map the latent factors across domains more accurately.

**Result 3: Cross-System Recommendation (Tasks 3 & 4)**

In order to answer question **Q2**, we compare the performances of our methods and the seven comparison methods in the CSR tasks (Tasks 3 & 4). From Table 2, we can see that, except when $K = 10$ in Task 4, BPR_DCDCSR outperforms MMMF_DCDCSR and PMF_DCDCSR because its MF model can create a large number of triples on the target system Douban* to train the parameters, which can generate relatively accurate latent factors. In terms of MAE, BPR_DCDCSR outperforms all the seven comparison methods by an average of 4.20%, ranging from 3.00% to 9.00%. Moreover, in terms of RMSE, BPR_DCDCSR outperforms all the seven comparison methods by an average of 4.46%, ranging from 3.43% to 9.08%. Compared to all the seven comparison methods, our methods perform clearly better because our sparsity guided DNN mapping process maps the latent factors across systems more accurately.

Furthermore, compared to Result 2, in Result 3, our methods deliver more improvements in terms of MAE and RMSE. This is because our methods can effectively utilize the ratings of the source systems MovieLens and Netflix when they are much richer than those of the target system Douban*.

**Summary:** According to Result 1, we can answer question **Q1** as follows: In general, the performances of DCDCSR methods increase with the dimension $K$ of the latent factors when $K \in \{10, 20\}$. However, when $K \in \{50, 100\}$, the performances have no significant improvement even declines slightly. According to Results 2 & 3, we can answer question **Q2** as follows: In general, our DCDCSR methods outperform all the comparison methods for both CDR and CSR because our sparsity guided DNN mapping process can map latent factors across domains or systems more accurately. In addition, the comparison of Results 2 & 3 demonstrates that our methods can effectively utilize more rating data.

## 5 Conclusions

In this paper, we have proposed a Deep framework for both CDR and CSR, called DCDCSR, which is based on MF models and a fully connected Deep Neural Network (DNN). The DNN is applied to more accurately map the latent factors across domains or systems. In addition, we utilized the sparsity degrees of individual users and items in the source and target domains or systems to guide the DNN training process, which can effectively utilize more rating data. The superior performances of our model have been demonstrated by extensive experiments conducted on three real-world datasets.

## Acknowledgements

This work was partially supported by Australian Research Council Discovery Project DP180102378.